\documentclass[journal,twoside,web]{ieeecolor}
\usepackage{generic}
\usepackage{cite}
\usepackage{amsmath,amssymb,amsfonts}
\usepackage{algorithmic}
\usepackage{graphicx}
\usepackage{textcomp}
\usepackage{url}

\usepackage{booktabs}
\usepackage{graphicx}
\usepackage{tablefootnote}
\usepackage{soul}
\usepackage{multirow}
\usepackage{caption}

\def\BibTeX{{\rm B\kern-.05em{\sc i\kern-.025em b}\kern-.08em
    T\kern-.1667em\lower.7ex\hbox{E}\kern-.125emX}}
\begin{document}
\title{Large AI Models in Health Informatics: Applications, Challenges, and the Future}
\author{Jianing Qiu, Lin Li, Jiankai Sun, Jiachuan Peng, Peilun Shi, \\Ruiyang Zhang, Yinzhao Dong, Kyle Lam, Frank P.-W. Lo, Bo Xiao, \\Wu Yuan,  \IEEEmembership{Senior Member, IEEE}, Ningli Wang, Dong Xu, and Benny Lo, \IEEEmembership{Senior Member, IEEE}
% \thanks{This paragraph of the first footnote will contain the date on 
% which you submitted your paper for review. It will also contain support 
% information, including sponsor and financial support acknowledgment. For 
% example, ``This work was supported in part by the U.S. Department of 
% Commerce under Grant BS123456.'' }
% \thanks{The next few paragraphs should contain 
% the authors' current affiliations, including current address and e-mail. For 
% example, F. A. Author is with the National Institute of Standards and 
% Technology, Boulder, CO 80305 USA (e-mail: author@boulder.nist.gov). }
\thanks{Jianing Qiu is with the Department of Computing, Imperial College London, U.K., and he is also with the Department of Biomedical Engineering, The Chinese University of Hong Kong, Hong Kong SAR. He was with Precision Robotics (Hong Kong) Ltd., Hong Kong SAR (e-mail: jianing.qiu17@imperial.ac.uk).}
\thanks{Lin Li is with the Department of Informatics, King's College London, U.K. (e-mail: lin.3.li@kcl.ac.uk).}
\thanks{Jiankai Sun is with the School of Engineering, Stanford University, USA (e-mail: jksun@stanford.edu).}
\thanks{Jiachuan Peng is with the Department of Engineering Science, University of Oxford, U.K. (e-mail: jiachuan.peng@seh.ox.ac.uk).}
\thanks{Peilun Shi and Wu Yuan are with the Department of Biomedical Engineering,  The Chinese University of Hong Kong, Hong Kong SAR (e-mails: peilunshi@cuhk.edu.hk and wyuan@cuhk.edu.hk).}
\thanks{Ruiyang Zhang is with Precision Robotics (Hong Kong) Ltd., Hong Kong SAR (e-mail: r.zhang@prhk.ltd).}
\thanks{Yinzhao Dong is with the Faculty of Engineering, The University of Hong Kong, Hong Kong SAR (e-mail: dongyz@connect.hku.hk).}
\thanks{Kyle Lam is with the Department of Surgery and Cancer, Imperial
College London, U.K. (e-mail: k.lam@imperial.ac.uk).}
\thanks{Frank P.-W. Lo  and Bo Xiao are with the Hamlyn Centre for Robotic Surgery, Imperial College London, U.K. (e-mails: po.lo15@imperial.ac.uk and b.xiao@imperial.ac.uk).}
\thanks{Ningli Wang is with Beijing Tongren Eye Center, Beijing Tongren Hospital, Capital Medical University, and also with Beijing Ophthalmology \& Visual Sciences Key Laboratory, Beijing, China (e-mail: wningli@vip.163.com)}
\thanks{Dong Xu is with the Department of Electrical Engineering and Computer Science, and the Christopher S. Bond Life Sciences Center, University of Missouri, USA (e-mail: xudong@missouri.edu).}
\thanks{Benny Lo is with the Facualty of Medicine, Imperial College London, U.K., and he is also with Precision Robotics (Hong Kong) Ltd., Hong Kong SAR (e-mail: benny.lo@imperial.ac.uk).}
\thanks{Corresponding authors: Wu Yuan and Benny Lo}
}

\maketitle

\begin{abstract}
Large AI models, or foundation models, are models recently emerging with massive scales both parameter-wise and data-wise, the magnitudes of which can reach beyond billions. Once pretrained, large AI models demonstrate impressive performance in various downstream tasks. A prime example is ChatGPT, whose capability has compelled people’s imagination about the far-reaching influence that large AI models can have and their potential to transform different domains of our lives. In health informatics, the advent of large AI models has brought new paradigms for the design of methodologies. The scale of multi-modal data in the biomedical and health domain has been ever-expanding especially since the community embraced the era of deep learning, which provides the ground to develop, validate, and advance large AI models for breakthroughs in health-related areas. This article presents a comprehensive review of large AI models, from background to their applications. We identify seven key sectors in which large AI models are applicable and might have substantial influence, including 1) bioinformatics; 2) medical diagnosis; 3) medical imaging; 4) medical informatics; 5) medical education; 6) public health; and 7) medical robotics. We examine their challenges, followed by a critical discussion about potential future directions and pitfalls of large AI models in transforming the field of health informatics.
\end{abstract}


\section{Introduction}\label{sec:introduction}

\begin{figure}[!t]
\centerline{\includegraphics[width=\linewidth]{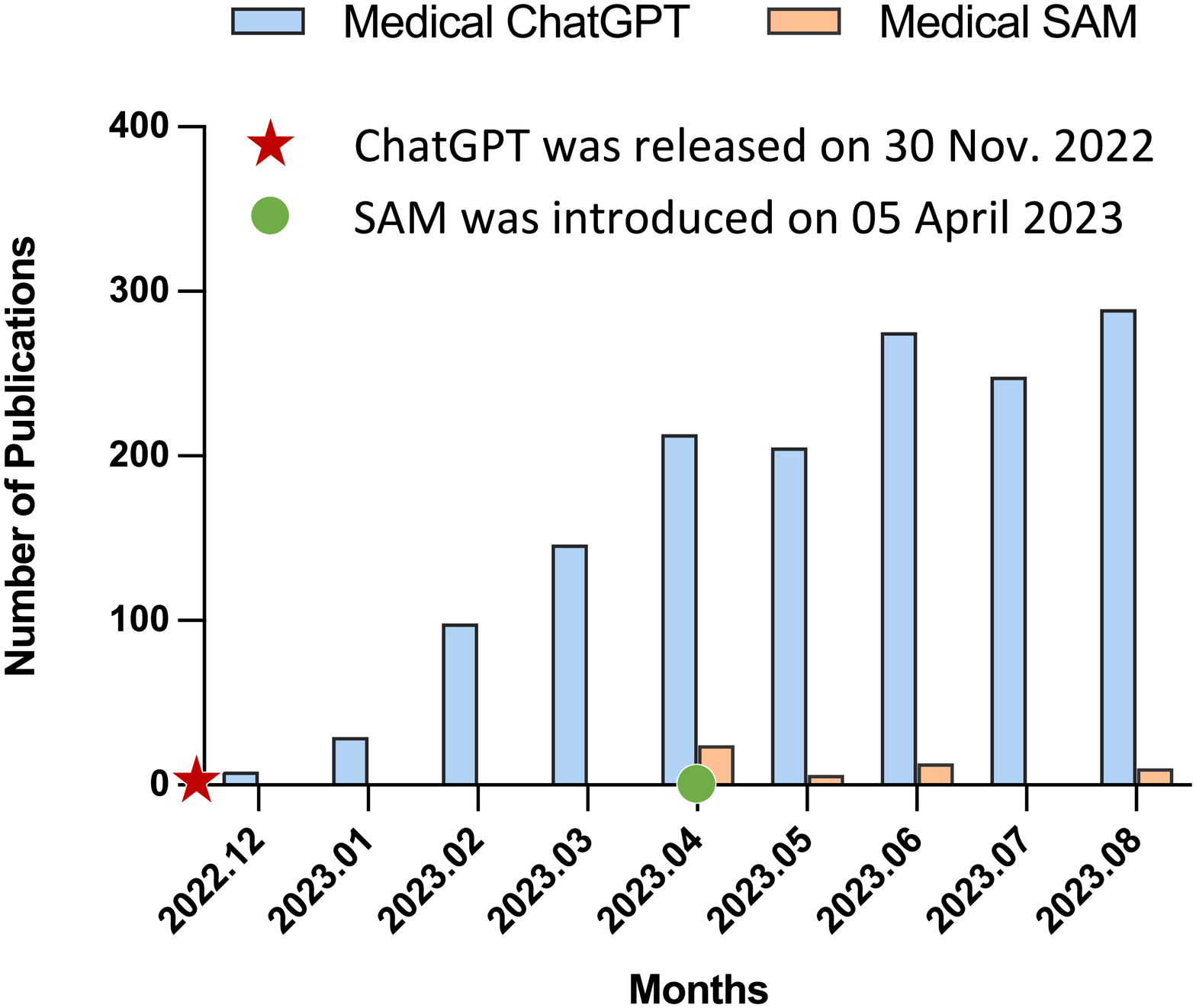}}
\caption{Number of publications related to ChatGPT and SAM in medical and health areas. Statistics were queried from Google Scholar with the keywords ``Medical ChatGPT" or ``Medical Segment Anything", and the last entry was 31-th Aug. 2023. From April to August, each month, there were over 200 publications about ChatGPT in medicine and healthcare.}
\label{fig:chatgpt_sam_publications}
\vspace{-0.5cm}
\end{figure}

\begin{figure*}[!t]
\centerline{\includegraphics[width=\textwidth]{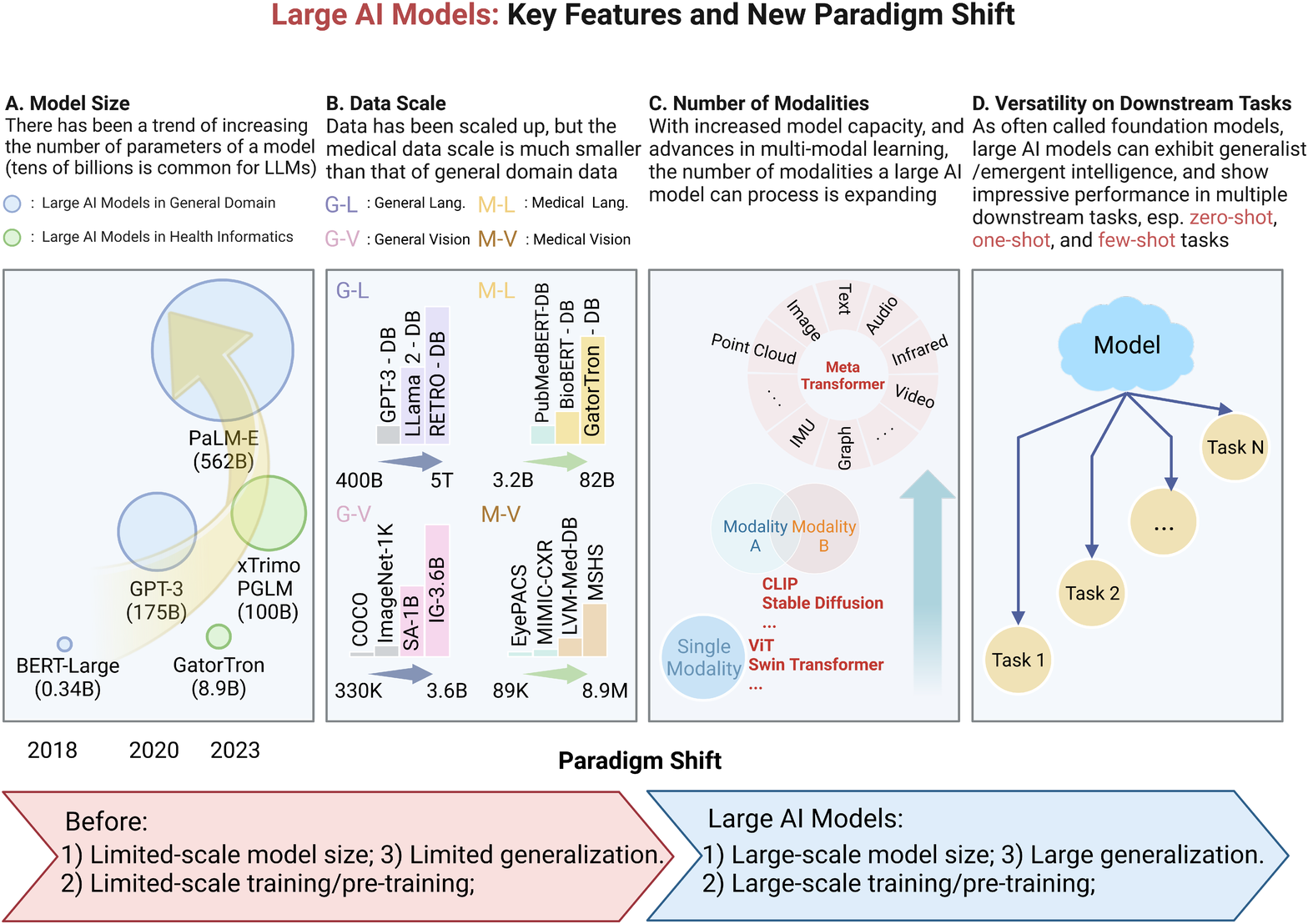}}
\caption{The key features of large AI models lie in the following four aspects: 1) increased size (e.g., for large language models (LLMs), the number of parameters is often billions); 2) trained with large-scale data (e.g., for LLMs, the data can contain trillions of tokens; and for large vision models (LVMs), the data can contain billions of images); 3) able to process data of multiple modalities; and 4) can perform well across multiple downstream tasks, especially on zero-, one-, and few-shot tasks.}
\label{fig:LAM_key_feature_summary}
\vspace{-0.5cm}
\end{figure*}

\IEEEPARstart{T}{he} introduction of ChatGPT~\cite{chatgpt2022} has triggered a new wave of development and deployment of Large AI Models (LAMs) recently. As shown in Fig.~\ref{fig:chatgpt_sam_publications}, ChatGPT and the phenomenal Segment Anything Model (SAM)~\cite{kirillov2023segment} have sparked active research in medical and health sectors since their initial launch. Although groundbreaking, the AI community has in fact started creating LAMs much earlier, and it was the seminal work introducing the Transformer model~\cite{vaswani2017attention} back in 2017 that accelerated the creation of LAMs.

The recent advances in data science and AI algorithms have endowed LAMs with strengthened \textit{generative} and \textit{reasoning} capabilities, as well as \textit{generalist intelligence} across multiple tasks with impressive zero- and few-shot performance, significantly distinguishing them from early deep models. For example, when asked for medical advice, ChatGPT, based on GPT-4~\cite{gpt4openai2023}, demonstrates the capability of recalling prior conversation and being able to contextualize the user's past medical history before answering, showing a new level of intelligence way beyond that of a simple symptom checker~\cite{lee2023ai}.

One notable bottleneck of developing supervised medical and clinical AI models is that they require annotated data at scale for training a well-functioning model. However, such annotations have to be conducted by domain experts, which is often expensive and time-consuming. This causes the curation of large-scale medical and clinical data with high-quality annotations to be challenging. However, this may no longer be a bottleneck for LAMs, as they can leverage self-supervision and reinforcement learning in training, relieving the annotation burden and workload of curating large-scale annotated datasets~\cite{gulshan2016development}. With the ever-increasing proliferation of medical Internet of things such as pervasive wearable sensors, medical and clinical history such as electronic health records (EHRs), prevalent medical imaging for diagnosis such as computed-tomography (CT) scans, the growing genomic sequence discovery, and more, the abundance of biomedical, clinical, and health data fosters the development of the next generation of AI models in the field, which are expected to have a large capacity for modeling the complexity and magnitude of health-related data, and generalize to multiple unseen scenarios to actively assist and engage in clinical and medical decision-making.

Despite the homogeneity of the model architecture (current LAMs are primarily based on Transformer~\cite{vaswani2017attention}), LAMs inherently are strong learners of heterogeneous data due to their large capacity, unified input modeling of different modalities, and improved multi-modal learning techniques. Multi-modality is common in biomedical and health settings, and the multi-modal nature of health data provides the natural and promising ground for developing and evaluating LAMs.

The LAMs that this article discusses are mainly foundation models~\cite{bommasani_opportunities_2022}. However, this article also provides a retrospective of the recent LAMs that are not necessarily considered foundational at their current stage, but are seminal in advancing the future development of LAMs in the fields of biomedicine and health informatics. Fig.~\ref{fig:LAM_key_feature_summary} summarizes the key features of LAMs, and highlights the paradigm shift it is introducing, i.e., 1) large-scale model size; 2) large-scale training/pre-training; and 3) large generalization.

Albeit inspirational, LAMs still face challenges and limitations, and the rapid rise of LAMs brings new opportunities as well as potential pitfalls. This article aims to provide a comprehensive review of the recent developments of LAMs, with a particular focus on their impacts on the biomedical and health informatics communities. The remainder of this article is organized as follows: Section~\ref{sec:background_large_ai_model} describes the background of LAMs in general domains, such as natural language processing (NLP) and computer vision (CV); Section~\ref{sec:applications} discusses current progress and possible applications of LAMs in key sectors of health informatics; Section~\ref{sec:challenges} discusses challenges, limitations and risks of LAMs; Section~\ref{sec:future} points out some potential future directions of advancing LAMs in health informatics, and Section~\ref{sec:conclusions} concludes.  

As this field progresses very rapidly, and also due to the page limit, there are a lot of works that this paper cannot cover. It is our hope that the community can be updated with the latest advances, so we refer readers to our website \footnote{https://github.com/Jianing-Qiu/Awesome-Healthcare-Foundation-Models} for the latest progress about LAMs.

\section{Background of Large AI Models}\label{sec:background_large_ai_model}

The burgeoning AI community has devoted much effort to developing large AI models (LAMs) in recent years by leveraging the massive influx of data and computational resources. Based on the pre-training data modality, this article categorizes the current LAMs into three types and defines them as follows:

\begin{enumerate}
    \item Large Language Model (LLM): LLMs are pre-trained on language data and applied to language downstream tasks. Language in different settings can have different interpretations, e.g., protein is the language of life, and code is the language of computers.
    \item Large Vision Model (LVM): LVMs are pre-trained on vision data and applied to vision downstream tasks.
    \item Large Multi-modal Model (LMM): LMMs are pre-trained on multi-modal data, e.g., language and vision data, and applied to various single- or multi-modal downstream tasks.
\end{enumerate}

This section provides an overview of the background of these three types of LAMs in general domains.

\subsection{Large Language Models}

The proposal of the Transformer architecture~\cite{vaswani2017attention} heralds the start of developing large language models (LLMs) in the field of NLP. Since 2018, following the birth of GPT (Generative Pre-trained Transformer)~\cite{radford2018improving} and BERT (Bidirectional Encoder Representations from Transformers)~\cite{devlin2018bert}, the development of LLMs has progressed rapidly.

Broadly speaking, the recent LLMs~\cite{touvron2023llama,touvron2023llama2,chung2022scaling,chowdhery2022palm,hoffmann2022training,smith2022using,scao2022bloom,thoppilan2022lamda,zhang2022opt,ouyang2022training,rae2021scaling,sanh2021multitask}, have the following three distinct characteristics: 1) parameter-wise, the number of learnable parameters of an LLM can be easily scaled up to billions; 2) data-wise, a large volume of unlabelled data are used to pre-train an LLM, and the amount can often reach millions or billions if not more; 3) paradigm-wise, LLMs are first pre-trained often with weakly- or self-supervised learning (e.g., masked language modeling~\cite{devlin2018bert} and next token prediction~\cite{gpt4openai2023}), and then fine-tuned or adapted to various downstream tasks such as question answering and dialogue in which they are able to demonstrate impressive performance. 

Recent advances reveal that LLMs are impressive zero-shot, one-shot, and few-shot learners. They are able to extract, summarize, translate, and generate textual information with only a few or even no prompt/fine-tuning samples~\cite{gpt4openai2023}. Furthermore, LLMs manifest impressive reasoning capability, and this capability can be further strengthened with prompt engineering techniques such as Chain-of-Thought prompting~\cite{wei2022chain}.

There was an upsurge in the number of new LLMs from 2022 onwards. Despite the general consensus that scaling up the number of parameters and the amount of data will lead to improved performance, which leads to a dominant trend of developing LLMs often with billions of parameters (e.g., LLMs such as PaLM~\cite{chowdhery2022palm} have already contained 540 billion parameters) and even trillions of data tokens (e.g., LLaMa 2 was pre-trained with 2 trillion tokens~\cite{touvron2023llama2}, and the training data of RETRO~\cite{borgeaud2022improving} had over 5 trillion tokens), there is currently no concerted agreement within the community that if this continuous growth of model and data size is optimal~\cite{hoffmann2022training,touvron2023llama}, and there is also lacking a verified universal scaling law.

To balance the data annotation cost and efficacy, as well as to train an LLM that can better align with human intent, researchers have commonly used reinforcement learning from human feedback (RLHF)~\cite{christiano2017deep} to develop LLMs that can exhibit desired behaviors. The core idea of RLHF is to use human preference datasets to train a Reward Model (RM), which can predict the reward function and be optimized by RL algorithms (e.g., Proximal Policy Optimization (PPO)~\cite{schulman2017proximal}). The framework of RLHF has attracted much attention and become a key component of many LLMs, such as InstructGPT~\cite{ouyang2022training}, Sparrow~\cite{glaese2022improving}, and ChatGPT~\cite{chatgpt2022}. Recently, Susano Pinto et al.~\cite{susano2023tuning} have also investigated this reward optimization in vision tasks, which can possibly advance the development of future LVMs using RLHF.

\subsection{Large Vision Models} \label{sec: background lvms}

In computer vision, it has been a common practice for years to first pre-train a model on a large-scale dataset and then fine-tune it on the dataset of interest (usually smaller than the one for pre-training) for improved generalization \cite{yosinski2014transferable}. The fundamental changes driving this evolution of large models lie in \textbf{the scale of pre-training datasets and models}, and \textbf{the pre-training methods}. ImageNet-1K (1.28M images) \cite{deng2009imagenet} and -21K (14M images) \cite{ridnik1imagenet} used to be canonical datasets for visual pre-training. ImageNet is manually curated for high-quality labeling, but the prohibitive cost of curation severely hindered further scaling. To push the scale beyond ImageNet's, datasets like JFT (300M \cite{sun2017revisiting} and 3B \cite{zhai2022scaling} images) and IG (3.5B \cite{mahajan2018exploring} and 3.6B \cite{singh2022revisiting} images) were collected from the web with less or no curation. The quality of annotation is therefore compromised, and the accessibility of datasets become limited because of copyright issues.

The compromised annotation requires the pre-training paradigm to shift from supervised learning to weakly-/self-supervised learning or unsupervised learning. The latter methods include autoregressive modeling, generative modeling and contrastive learning. Autoregressive modeling trains the model to autoregressively predict the next pixel conditioned on the preceding pixels \cite{chen_generative_2020}. Generative modeling trains the model to reconstruct the entire original image, or some target regions within it \cite{assran2023self}, from its corrupted \cite{chen2021pre} or masked \cite{he_masked_2022} variants. Contrastive learning trains the model to discriminate similar and/or dissimilar data instances \cite{chen2020simple}.

Vision Transformers (ViTs) and Convolutional Neural Networks (CNNs) are two major architectural families of LVMs. For vision transformers, pioneering works ViT \cite{dosovitskiy_image_2021} and iGPT \cite{chen_generative_2020} transferred the transformer architectures from NLP to CV with minimal modification, but the resulting architectures incur high computational complexity, which is quadratic to the image size. Later, works like TNT \cite{han_transformer_2021} and Swin Transformer \cite{liu2021swin} were proposed to better adapt transformers to visual data. Recently, ViT-G/14 \cite{zhai2022scaling}, SwinV2-G \cite{liu_swin_2022} and ViT-22B \cite{Dehghani2023ScalingVT} substantially scaled the vision transformers up using a bag of training tricks to achieve state-of-the-art (SOTA) accuracy on various benchmarks. While ViTs may seem to gain more momentum than CNNs in developing LVMs, to improve CNNs, the latest works such as ConvNeXt \cite{liu2022convnet} and InternImage \cite{wang2023internimage} redesigned CNN architecture with inspirations from ViTs and achieved SOTA accuracy on ImageNet. This refutes the previous statement that CNNs are inferior to ViTs. Apart from the above, recent works like CoAtNet \cite{dai2021coatnet} and ConViT \cite{d2021convit} merge CNNs and ViTs to form new hybrid architectures.
Note that ViT-22B is the largest vision model to date, whose scale is significantly larger than that (1.08B) of the current art of CNNs (InternImage) but is still much behind that of the contemporary LLMs.

Architecturally speaking, LVMs are largely-scaled-up variants of their base architectures. How they are scaled up can significantly impact the final performance. Simply increasing the depth by repeating layers vertically may be suboptimal \cite{kolesnikov2020big}, so a line of studies \cite{tan2019efficientnet, tan2021efficientnetv2, wang2023internimage} investigate the rules for effective scaling. Furthermore, scaling the model size up is usually combined with larger-scale pre-training \cite{kolesnikov2020big, goyal2021self} and efficient parallelism \cite{huang2019gpipe} for improved performance.

LVMs also transform other fundamental computer vision tasks beyond classification. The latest breakthrough in segmentation task is SAM \cite{kirillov2023segment}. SAM is built with a ViT-H image encoder (632M), a prompt encoder and a transformer-based mask decoder that predicts object masks from the output of the above two encoders.
Prompts can be points or bounding boxes in images or text. 
SAM demonstrates a remarkable zero-shot generalization ability to segment unseen objects and images. Furthermore, to train SAM, a largest segmentation dataset to date, SA-1B, with over 1B masks is constructed.

\subsection{Large Multi-modal Models} \label{sec: background lmm}

This section describes large multi-modal models (LMMs). While the primary focus is on one type of LMMs: large vision-language models (LVLMs), multi-modality beyond vision and language is also summarized in the end.

Training LVLMs like CLIP \cite{radford2021learning} requires more than hundreds of millions of image-text pairs. Such large amount of data was often closed source \cite{radford2021learning, jia_scaling_2021, yuan_florence_2021}. Until recently, LAION-5B \cite{schuhmann_laion-5b_2022} was created with 5.85B data samples, matching the size of the largest private dataset while being available to the public.

LVLMs usually adopt a dual-stream architecture: input text and image are processed separately by their respective encoders to extract features. For representation learning, the features from different modalities are then aligned through contrastive learning \cite{radford2021learning, jia_scaling_2021, yuan_florence_2021} or fused into a unified representation through another encoder on the top of all extracted features \cite{singh_flava_2022, wang_simvlm_2022, chen2022pali, li_blip-2_2023, huang_language_2023}. Typically, the entire model, including unimodal encoder and multi-modal encoder if have, is pre-trained on the aforementioned large-scale image-text datasets, and fine-tuned on the downstream tasks or to carry out zero-shot tasks without fine-tuning. The pre-training objectives can be multi-modal tasks only or with unimodal tasks (see Section~\ref{sec: background lvms}). Common multi-modal pre-training tasks contain 
image-text contrastive learning \cite{radford2021learning, jia_scaling_2021, yuan_florence_2021}, image-text matching \cite{singh_flava_2022, li_blip_2022, li_blip-2_2023}, autoregressive modeling \cite{huang_language_2023, wang_simvlm_2022}, masked modeling \cite{singh_flava_2022}, image-grounded text generation \cite{li_blip_2022, li_blip-2_2023}, etc. Recent studies suggest that scaling the unimodal encoders up \cite{pham2021combined, chen2022pali} and pre-training with multiple objectives across uni- and multi-modalities \cite{li_blip_2022, li_blip-2_2023} can substantially benefit multi-modal representation learning.

Recently, LVLMs made a major breakthrough in text-to-image generation. There are generally two classes of methods for such a task: autoregressive model \cite{ramesh_zero-shot_2021, yuscaling, ramesh2022hierarchical} and diffusion model \cite{rombach2021highresolution, ramesh2022hierarchical, nichol2022glide, saharia2022photorealistic}. Autoregressive model, like introduced in Section~\ref{sec: background lvms}, first concatenates the tokens (returned by some encoders) of text and images together and then learns a model to predict the next item in the sequence. In contrast, diffusion model first perturbs an image with random noise progressively until the image becomes complete noise (forward diffusion process) and then learns a model to gradually denoise the completely noisy image to restore the original image (reverse diffusion process) \cite{sohl2015deep}. Text description is first encoded by a separate encoder and then integrated into the reverse diffusion process as the input to the model so that the image generation can be conditioned on the text prompt. It is common to reuse those pre-defined LLM and LVM architectures and/or their pre-trained parameters as the aforementioned encoders. The scale of these encoders and the generator can significantly impact the quality of generation and the ability of language understanding \cite{yuscaling, saharia2022photorealistic}.

The paradigm of bridging language and vision modalities can be beyond learning, e.g., using LLM to instruct other LVMs to perform vision-language tasks \cite{wu2023visual}. Beyond vision and language, recent development in LMMs seeks to unify more modalities under one single framework, e.g., ImageBind~\cite{girdhar2023imagebind} combines six whereas Meta-Transformer~\cite{zhang2023meta} unifies twelve modalities.

\section{Applications of Large AI Models in Health Informatics}\label{sec:applications}

\begin{figure*}[!t]
\centering
\captionof{table}{Comparison between state-of-the-art LAMs (second row) and prior arts (first row) in typical tasks of seven biomedical and health sectors. ``$-$'' denotes not applicable. ``N/R" denotes not released in the original literature. For zero-shot medical segmentation task, we tested the off-the-shelf Mask RCNN model to compare with SAM.}
\centerline{\includegraphics[width=\textwidth]{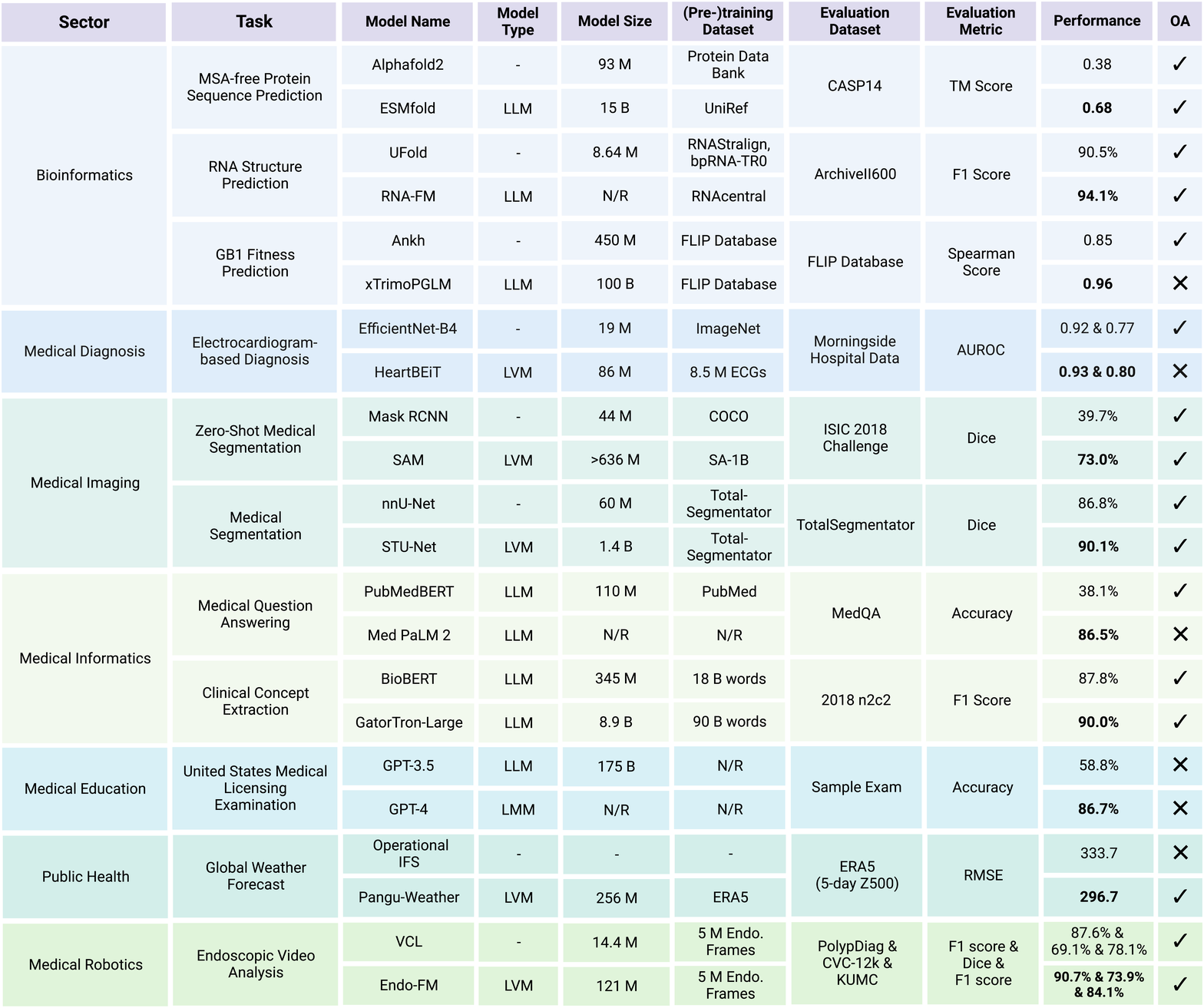}}
\label{tab:LAM_sota}
\vspace{-0.5cm}
\end{figure*}

In this section, we identify seven key sectors in which LAMs will have substantial influence and bring a new paradigm for tackling the problems and challenges in health informatics. The seven key sectors include 1) bioinformatics; 2) medical diagnosis; 3) medical imaging; 4) medical informatics; 5) medical education; 6) public health; and 7) medical robotics. Table~\ref{tab:LAM_sota} compares current LAMs with previous SOTA methods in these seven sectors.

\subsection{Bioinformatics}\label{subsec:mole_bio_drug_disconvery}

Molecular biology studies the roles of biological macromolecules (e.g., DNA, RNA, protein) in life processes and describes various life activities and phenomena including the structure, function, and synthesis of molecules. Although many experimental attempts have been made on this topic over decades~\cite{bai2015cryo,wuthrich2001way,grimes2018crystallography}, they are still of high cost, long experiment cycle, and high production difficulty. For example, the number of experimentally determined protein structures stored in the protein data bank (PDB) hardly rivals the number of protein sequences that have been generated. Efficient and accurate computational methods are therefore needed and can be used to accelerate the protein structure determination process. Due to the huge number of parameters and learning capacity, LAMs endow us with prospects to approach such a Herculean task. Especially, LLMs' outstanding representation learning ability has been employed to implicitly model the biological properties hidden in large-scale unlabeled data including RNA and protein sequences.

When it comes to the field of protein, starting from amino acid sequences, we can analyze the spatial structure of proteins and furthermore understand their functions, and mutual interactions. AlphaFold2~\cite{jumper2021highly} pioneered leveraging the attention-based Transformer model~\cite{vaswani2017attention} to predict protein structures. Specifically, they treated structure prediction as a 3D graph inference problem, where the network's inputs are pairwise features between residues, available templates, and multi-sequence alignment (MSA) embeddings. Especially, embeddings extracted from MSA can infer the evolutionary information between aligned sequences. Evoformer and structure modules were proposed to update the input representation and predict the final 3D structure, the whole process of which was recycled several times. Meanwhile, despite being trained on single-protein chains, AlphaFold2 exhibits the ability to predict multimers. To further enable multimeric inputs for training, DeepMind proposed AlphaFold-Multimer\cite{evans2021protein}, achieving impressive performance, especially in heteromeric protein complexes structure prediction.  Specifically, positional encoding was improved to encode chains, and multi-chain MSAs were paired based on species annotations and target sequence similarity.

In spite of the groundbreaking endeavors aforementioned works have contributed, to achieving optimal prediction, they still heavily rely on MSAs and templates searched from genetic and structure databases, which is time-consuming. Analogous to mining semantic information in natural language, researchers managed to explore co-evolution information in protein sequences in a self-supervised manner by employing large-scale protein language models (PLMs), which learn the global relation and long-range dependencies of unaligned and unlabelled protein sequences. ProGen (1.2B)\cite{madani2020progen} utilized a conditional language model to provide controllable generation of protein sequences. By inputting desired tags (e.g., function, organism), ProGen can generate corresponding proteins such as enzymes with good functional activity. Elnaggar et al.\cite{article} devised ProtT5-XXL (11B) which was first trained on BFD\cite{steinegger2018clustering} and then fine-tuned on UniRef50\cite{suzek2015uniref} to predict the secondary structure. ESMfold\cite{lin2023evolutionary} scaled the number of model parameters up to 15B and observed a significant prediction improvement over AlphaFold2 (0.68 vs 0.38 for TM-score on CASP14) with considerably faster inference speed when MSAs and templates are unavailable. Similarly, from only the primary sequence input, OmegaFold\cite{wu2022high} can outperform MSA-based methods~\cite{jumper2021highly,baek2021accurate}, especially when predicting orphan proteins that are characterized by the paucity of homologous structure. xTrimoPGLM\cite{Chen2023.07.05.547496} proposed a unified pre-training strategy that integrates the protein understanding and generation by optimizing masked language modelling and general language modelling concurrently and achieved remarkable performance over 13 diverse protein tasks with its 100B parameters. For instance, for GB1 fitness prediction in protein function task, xTrimoPGLM outperforms the previous SOTA method: Ankh\cite{elnaggar2023ankh}, with an 11\% performance increase. Moreover, for antibody structure prediction, xTrimoPGLM outperformed AlphaFold2 (TM-score: 0.951) and achieved SOTA performance (TM-score: 0.961) with significantly faster inference speed. We underscore that in the presence of MSA, although the performance of PLMs is hardly on par with Alphfold2, PLMs can make predictions several orders of magnitude faster, which speeds up the process of related applications such as drug discovery. In addition, because PLMs implicitly understand the deep information implied in protein sequences, they are promising to predict mutations in protein structures and their potential impact to help guide the design of next-generation vaccines.

In the context of RNA structure prediction, the number of nonredundant 3D RNA structures stored in PDB is significantly less than that of protein structures, which hinders the accurate and generalizable prediction of RNA structure from sequence information using deep learning. To mitigate the severe unavailability of labeled RNA data, Chen et al.\cite{chen2022interpretable} proposed the RNA foundation model (RNA-FM), which learns evolutionary information implicitly from 23 million unlabeled ncRNA sequences\cite{rnacentral2021rnacentral} by recovering masked nucleotide tokens, to facilitate multiple downstream tasks including RNA secondary structure prediction and 3D closeness prediction. Especially, for secondary structure prediction, RNA-FM achieves 3-5\% performance increase among three metrics (i.e., Precision, Recall, and F1-score), compared to UFold\cite{fu2022ufold} which utilizes U-Net as the backbone. Furthermore, based on RNA-FM, Shen et al.\cite{shen2022e2efold} pioneered predicting 3D RNA structure directly.

Undoubtedly, these models are seminal and have reduced the time and cost of molecule structure prediction by a large margin. Thereby, this raises the question of whether LAMs can completely replace experimental methods such as Cryo-EM\cite{bai2015cryo}. We deem that it still falls short from that point. Specifically, the advance of LAMs builds upon big data and large model capacity, which means they are still data-driven, and hence their ability to predict unseen types of data could sometimes be problematic. For instance, \cite{buel2022can} stated that AlphaFold can barely handle missense mutation on protein structure due to the lack of a corresponding dataset. Furthermore, how we can assess the quality of model prediction for unknown protein structures remains unclear. In turn, these unverified protein structures cannot be applied to, for example, drug discovery. Therefore, protocols and metrics need to be established to assess their quality and potential impacts. There are mutual and complementary benefits between LAMs and conventional experimental techniques. LAMs can be re-designed to predict the process of protein folding and reveal their mutual interactions so as to facilitate experimental methods. On the other hand, experimental information, such as some physical properties of molecules, can be leveraged by LAMs to further improve prediction performance, especially when dealing with rare data (e.g., orphan protein).

\subsection{Medical Diagnosis}

As research has been carried out to improve the safety and strengthen the factual grounding of LAMs, it is foreseeable that LAMs will play a significant role in medical diagnosis and decision-making.

CheXzero~\cite{tiu2022expert}, a zero-shot chest X-ray classifier, has demonstrated radiologist-level performance in classifying multiple pathologies which it never saw in its self-supervised learning process. Recently, ChatCAD~\cite{wang2023chatcad}, a framework that integrates multiple diagnostic networks with ChatGPT, demonstrated a potential use case for applying LLMs in computer-aided diagnosis (CAD) for medical images. By stratifying the decision-making process with specialized medical networks, and followed by an iteration of prompts based on the outcomes of those networks as the queries to an LLM for medical recommendations, the workflow of ChatCAD offers an insight into the integration of the LLMs that were pre-trained using a massive corpus, with the upstream specialized diagnostic networks for supporting medical diagnosis and decision-making. Its follow-up work ChatCAD+~\cite{zhao2023chatcad+}, shows improved quality of generating diagnostic reports with the incorporation of a retrieval system. Using external knowledge and information retrieval can potentially enable the resulting diagnostics more factually-grounded, and such a design has also been favoured and implemented in the ChatDoctor model~\cite{li2023chatdoctor}. By leveraging a linear transformation layer to align two medical LAMs, XrayGPT~\cite{thawkar2023xraygpt}, a conversational chest X-ray diagnostic tool, shows decent accuracy in responding to diagnostic summary. While most LLMs are based on English, researchers have also managed to fine-tune LLaMa~\cite{touvron2023llama}, an LLM, with Chinese medical knowledge, and the resulting model shows improved medical expertise in Chinese~\cite{wang2023huatuo}.

Apart from chest X-ray diagnostics and medical question answering, LAMs have also been applied to other diagnostic scenarios. HeartBEiT~\cite{vaid2023foundational}, a foundation model pre-trained using 8.5 million electrocardiograms (ECGs), shows that large-scale ECG pre-training could produce accurate cardiac diagnosis and improved explainability of the diagnostic outcome, and the amount of annotated data for downstream fine-tuning could be reduced. Medical LAMs may also potentially produce a more reliable forecast of treatment outcomes and the future development of diseases using their strong reasoning capability. For example, Li et al.~\cite{li2020behrt} proposed BEHRT, which is able to predict the most likely disease of a patient in his/her next visit by learning from a large archive of EHRs. Rasmy et al.~\cite{rasmy2021medbert} proposed Med-BERT, which is able to predict the heart failure of diabetic patients.

With the ubiquity of internet, medical LAMs can also offer remote diagnosis and medical consultation for people at home, providing people in need with more flexibility. We also envision that future diagnosis of complex diseases may also be conducted or assisted by a panel of clinical LAMs.

\subsection{Medical Imaging}

The adoption of medical imaging and vision techniques has vastly influenced the process of diagnosis and treatment of a patient. The wide use of medical imaging, such as CT and MRI, has produced a vast amount of multi-modal, multi-source, and multi-organ medical vision data to accelerate the development of medical vision LAMs.

The recent success of SAM~\cite{kirillov2023segment} has drawn much attention within the medical imaging community. SAM has been extensively examined in medical imaging, especially on its zero-shot segmentation ability. While research revealed that for certain medical imaging modalities and targets, the zero-shot performance of SAM is impressive (e.g., on endoscopic and dermoscopic images, as these are essentially RGB images, which are the same type as that of SAM's pre-training images), for imaging modalities that are medicine-specific such as MRI and OCT (optical coherence tomograpy), SAM often fails to segment targets in a zero-shot way~\cite{shi2023generalist}, mainly because the topology and presentation of a target in those imaging modalities are much different from what SAM has seen during pre-training. Nevertheless, after adaptation and fine-tuning, the medical segmentation accuracy of SAM can surpass current SOTA with a clear margin~\cite{wu2023medical}, showing the potential of extending versatility of general LVMs to medical imaging with parameter-efficient adaptation. Apart from zero-shot segmentation, MedCLIP~\cite{wang2022medclip} was proposed, a contrastive learning framework for decoupled medical images and text, which demonstrated impressive zero-shot medical image classification accuracy. In particular, it yielded over 80\% accuracy in detecting Covid-19 infection in a zero-shot setting. The recent PLIP model~\cite{huang2023visual}, built using image-text pairs curated from medical Twitter, enables both image-based and text-based pathology image retrieval, as well as enhanced zero-shot pathology image classification compared to CLIP~\cite{radford2021learning}.

Many medical imaging modalities are 3-dimensional (3D), and thus developing 3D medical LVMs are crucial. Med3D~\cite{chen2019med3d}, a heterogeneous 3D framework that enables pre-training on multi-domain medical vision datasets, shows strong generalization capabilities in downstream tasks, such as lung segmentation and pulmonary nodule classification.

With the success of generative LAMs such as Stable Diffusion~\cite{rombach2021highresolution} in the general domain, which can generate realistic high-fidelity images with text descriptions, Chambon et al.~\cite{chambon2022adapting} recently fine-tuned Stable Diffusion on medical data to generate synthetic chest X-ray images based on clinical descriptions. The encouraging generative capability of Stable Diffusion in the medical domain may inspire more future research on using generative LAMs to augment medical data that are conventionally hard to obtain, and expensive to annotate.

Nevertheless, some compromises are also evident in medical vision LAMs. For example, the currently common practice of training LVMs and LMMs often limits the size of the medical images to shorten the training time and reduce the computational costs. The reduced size inevitably causes information loss, e.g., some small lesions that are critical for accurate recognition might be removed in a compressed downsampled medical image, whereas doctors could examine the original high-resolution image and spot these early-stage tumors. This may cause performance discrepancies between current medical vision LAMs and well-trained doctors. In addition, although research has shown that increasing medical LAM size and data size could improve medical domain performance of the model, e.g., STU-Net~\cite{huang2023stu}, a medical segmentation model with 1.4 billion parameters, the best practice of model-data scaling is yet to be conclusive in medical imaging and vision.

\subsection{Medical Informatics}

In medical informatics, it has been a topic of long-standing interest to leverage large-scale medical information and signals to create AI models that can recognize, summarize, and generate medical and clinical content. 

Over the past few years, with advances in the development of LLMs~\cite{devlin2018bert, brown2020language, chowdhery2022palm}, and the abundance of EHRs as well as public medical text outlets such as PubMed~\cite{pubmedabstract2023,pubmedcentral2023}, research has been carried out to design and propose Biomedical LLMs. Since the introduction of BioBERT~\cite{lee2020biobert}, a seminal Biomedical LLM which outperformed previous SOTA methods on various biomedical text mining tasks such as biomedical named entity recognition, many different Biomedical LLMs that stem from their general LLM counterparts have been proposed, including ClinicalBERT~\cite{alsentzer2019publicly}, BioMegatron~\cite{shin2020biomegatron}, BioMedRoBERTa~\cite{gururangan2020don}, Med-BERT~\cite{rasmy2021med}, BioELECTRA~\cite{raj2021bioelectra}, PubMedBERT~\cite{gu2021domain}, BioLinkBERT~\cite{yasunaga2022linkbert}, BioGPT~\cite{luo2022biogpt}, and Med-PaLM~\cite{singhal2022large}. 

The recent GatorTron~\cite{yang2022large} model (8.9 billion parameters) pre-trained with de-identified clinical text (82 billion words) revealed that scaling up the size of clinical LLMs leads to improvements on different medical language tasks, and the improvements are more substantial for complex ones, such as medical question answering and inference. Previously, the PubMedBERT work~\cite{gu2021domain} also suggested that pre-training an LLM with biomedical corpora from scratch can lead to better results than continually training an LLM that has been pre-trained on the general-domain corpora. While training large number of parameters may seem daunting, parameter-efficient adaptation techniques such as low-rank adaptation (LoRA)~\cite{hu2021lora} have enabled researchers to efficiently adapt a 13 billion LLaMa model to produce decent US Medical Licensing Exam (USMLE) answers, and the performance of a collection of such fine-tuned models, called MedAlpaca~\cite{han2023medalpaca} also reveals that increasing model size and quality of data can improve model's medical domain expertise. As LLMs start to show emergent abilities~\cite{wei2022emergent} with their size scaled up increasingly, Agrawal et al.~\cite{agrawal2022large} revealed that recent LLMs such as InstructGPT~\cite{ouyang2022training} and GPT-3~\cite{brown2020language} can well extract clinical information in a few-shot setting despite being not explicitly trained for the clinical domain. Med-PaLM~\cite{singhal2022large}, a Biomedical LLM with 540 billion parameters generated by applying instruction prompt tuning on Flan-PaLM~\cite{chung2022scaling} (which exhibited SOTA accuracy on MultiMedQA~\cite{singhal2022large}), demonstrated the ability to answer consumer medical questions that are comparable to the performance of clinicians. Its follow-up work, Med-PaLM 2~\cite{singhal2023towards}, further strengthens medical reasoning, and as shown in Table~\ref{tab:LAM_sota}, it has reached an accuracy of 86.5\% on the MedQA benchmark. As prompt engineering has become a key technique for investigating and improving LLMs, Li{\'e}vin et al.~\cite{lievin2022can} have also applied various prompt engineering on the GPT-3.5 series such as InstructGPT~\cite{ouyang2022training} to understand their abilities on medical question answering, and their results suggested that increasing Chain-of-Thoughts (CoTs)~\cite{wei2022chain} per question can deliver better, more interpretable medical question responses.

The impressive performance of Biomedical LLMs on medical language tasks shows their potential to be used to assist clinicians in processing, interpreting, and analyzing clinical and medical data more efficiently, and also to vastly reduce the time that clinicians have to spend on documenting EHRs.  Patel and Lam~\cite{patel2023chatgpt} recently shed insight on using ChatGPT~\cite{chatgpt2022} to generate discharge summaries, which could potentially relieve doctors from laborious writing and improve their clinical productivity. Biomedical LLMs can also assist in the writing of prior authorizations for insurance purposes, accelerating treatment authorizations~\cite{priorauth2023}. On the patient side, the zero-, one-, and few-shot learning capability of LLMs may enable them to provide personalized medical assistance based on the medical history of each individual patient. In addition, LLMs may also find them applicable in clinical trial matching. Based on candidates' demographics and medical history, a Biomedical LLM may effectively generate eligible matching, which accelerates clinical trial recruitment and initiation.

\subsection{Medical Education}

It is likely that future medical education will also be influenced by LAMs, as research continues to strengthen their scientific grounding and creative generation. Many LAMs, such as GPT-4~\cite{gpt4openai2023} and Med PaLM 2~\cite{singhal2023towards}, have already passed USMLE with a score of over 86\%, demonstrating sound knowledge spectrum and reasonable capabilities in bioethics, clinical reasoning, and medical management.

The generative capability of such LAMs may augment medical student learning and help them gain additional insights from AI-generated content as recently pointed out in~\cite{kung2023performance}. A LAM with wide knowledge and social compliance can act as a companion learning assistant, answering medical questions promptly and explaining intricate terms and practices in simple sentences. For example, the recent GPT-4 model~\cite{gpt4openai2023} can act as a Socratic tutor, leading a student step-by-step to find the answers by themselves, which is an important step towards practical adoption of LAMs in education as they can be steered to teach/assist students in a desired manner. The OPTICAL model proposed by Shue et al.~\cite{shue2023empowering} recently shows the feasibility of using LLMs to guide beginners in analyzing bioinformatics data. The sentence paraphrasing abilities of LLMs~\cite{dai2023chataug} such as ChatGPT may also help students with dyslexia in their learning. However, concerns about the illegitimate uses of LAMs such as plagiarism are practical and should raise awareness. A pilot study conducted by Mitchell et al.~\cite{mitchell2023detectgpt} proposed a zero-shot detector named DetectGPT, which is able to distinguish human-written or LLM-generated text. This attempt may lead to more research into developing reliable tools for verifying the content source and potentially countering the side effects of LAMs in education. 

For medical education givers, LAMs can potentially create novel teaching and exam contents, and diversify the teaching formats and their presentation. Based on the history of medical study and outcomes, LAMs may also help design personalized and precise course materials for students in need. In addition, LAMs may also help deliver remote medical education, providing engaging learning experiences and opportunities for students living in resource-poor areas or from underprivileged families. LAMs can also serve as a grading and scoring system in medical education, e.g., grading the surgical skill of a surgeon operating a surgical robot.

In medical and clinical training such as nurse training, one can imagine a domain-knowledgeable LAM can act as an assistant or a trainer to supervise the training. For certain frequent and tedious routine medical training courses, human trainers tend to become less productive as training keeps repeating, and the quality of training delivery also varies among different human trainers. With a wide knowledge spectrum and responsive interactions, training delivered by a LAM can potentially be more engaging and productive, and the standard of training can be maintained as equal and of high quality.

\begin{figure*}[!t]
\centering
\captionof{table}{Large-scale datasets in biomedical and health informatics}
\centerline{\includegraphics[width=\textwidth]{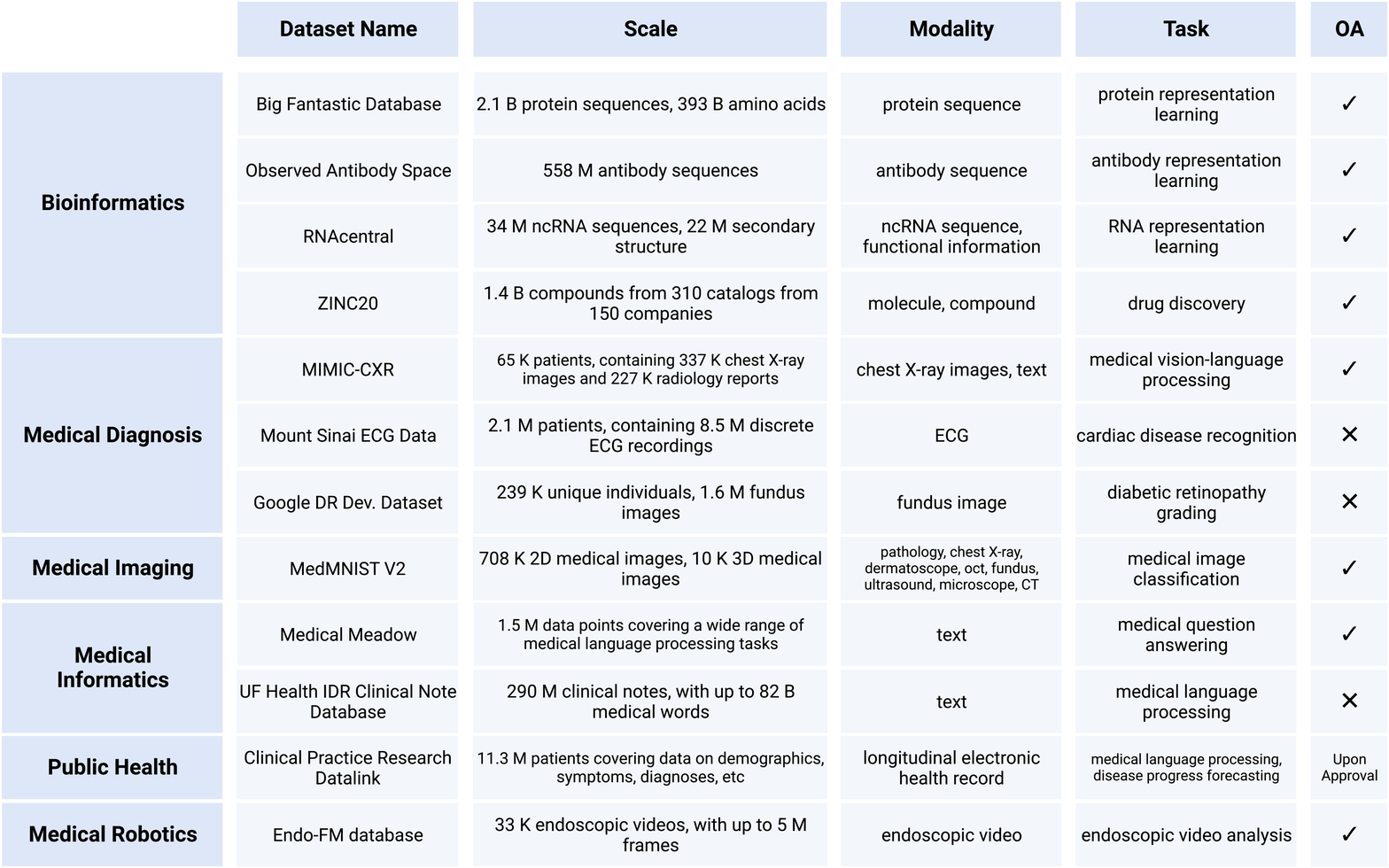}}
\label{tab:LAM_dataset}
\vspace{-0.5cm}
\end{figure*}

\subsection{Public Health}

As the American epidemiologist Larry Brilliant said \textit{``outbreaks are inevitable, but pandemics are optional"}, with the world gradually returning to normal after the Covid-19 pandemic, if there is one thing that the world has to reflect on, it is how we become prepared to prevent the next pandemic.

Based on past public policy and interventions to contain the spread of infectious diseases and the specific current situation, LLMs may help epidemiologists and policymakers to draft targeted public policies and recommend effective interventions. LLMs and other LAMs are also likely to be used to monitor, track, forecast, and analyze the progress of new outbreaks. LAMs have been actively researched for drug discovery, e.g., the Pangu Drug model~\cite{lin2022pangu}, and they can potentially be used for the design of vaccine and drugs to treat and save people from new outbreaks. Furthermore, another potential usage of LAMs, as pointed out in~\cite{korngiebel2021considering}, can be in precision triage and diagnosis, in which they could play a pivotal role as medical care workforce might be stretched when encountering a new outbreak. An important aspect of tackling an outbreak/epidemic is to handle misinformation. The study conducted by Chen et al.~\cite{chen2020tracking} revealed that from 21 January 2020 to 21 March 2020, Twitter produced over 72 million Covid-19-related tweets. If unverified media information proliferates at scale, it inevitably causes complications in tackling the outbreak. Although LAMs could be double-edged swords when it comes to misinformation, with gradually complete regulations and strengthened factual grounding of LAMs, they can be used to effectively identify misinformation and tackle public health infodemic.

Beyond their promising usage in preventing pandemics, LAMs are also an effective tool for solving other public health challenges, for example, providing large-scale dietary monitoring and assessment~\cite{peng2022clustering,qiu2023egocentric} to tackle the growing \textit{double burden of malnutrition}~\cite{popkin2019dynamics} in many low- and middle-income countries, and demystifying and proposing new solutions for mental illnesses that are common in populations. Researchers have recently proposed ClimaX~\cite{nguyen2023climax}, a foundation model for forecasting weather and climate change. With their remarkable forecasting capability, LAMs like ClimaX and Pangu-Weather~\cite{bi2023accurate} can advance our understanding of climate change and provide solutions to better address the global health issues posed by climate change.

\subsection{Medical Robotics}

From surgical robots that allow surgeons to perform precision minimally invasive surgery, to wearable robots that assist patients with health monitoring and rehabilitation, medical robotics has seen rapid growth and advances over the past few decades. LAMs have begun to show exciting prospects in enhancing medical robotic vision, interaction, and autonomy.

\subsubsection{Enhance Vision}

The integration of LAMs into surgical robots has the potential to enhance the vision of these systems in surgery. Endo-FM~\cite{wang2023foundation}, a foundation model with high precision for endoscopic video classification, segmentation, and detection, could be one of these LAMs to provide robotic surgery systems with enhanced vision.
In addition to online vision enhancement, LAMs can also potentially improve the offline workflow analysis of robotic surgery, and more accurately and objectively predict the likelihood of complications and successful outcomes, which help surgeons better plan and execute surgeries in the future. Furthermore, with their strong generative capabilities, LAMs can be used to generate and simulate surgical procedures, allowing surgeons to practice and refine their techniques before operating on a patient with real surgical robots. Beyond surgical robots, the perception of many companion and assistive robots can also be enhanced by LAMs, e.g., enabling a companion robot to better understand a patient's emotion through accurate recognition of facial expressions~\cite{d2022emotion}, and enabling an assistive robot to offer safer, more natural navigation for visually impaired people~\cite{qiu2022egocentric}.

\subsubsection{Improve Interaction}
LAMs may significantly improve the interactive capabilities of many medical robots, by enabling them to recognize human emotions, gestures, and speech, and respond to high-level human language commands. For example, this will be easier for patients undergoing rehabilitation to communicate and engage with their robotic assistants, improving their overall recovery experience. More intelligent LAMs may also better understand human intentions and create more human-like companionship, which could improve the overall quality of care for the elderly~\cite{asgharian2022review}. Recently, SurgicalGPT~\cite{seenivasan2023surgicalgpt}, a visual question answering model for surgery, has shown great promise that future robotic surgery could become more interactive between surgeons and the surgical robots.

\subsubsection{Increase Autonomy}

LAMs have the potential to turn robotic pipelines from the current \textit{engineer in the loop} to \textit{user in the loop} using high-level language commands~\cite{vemprala2023chatgpt}, which could enable surgeons with less programming proficiency to easily adapt robotic manipulations to their target tasks. Studies have proposed to use a single LAM to conduct diverse robotic tasks, demonstrating impressive adaptability and generalization skills~\cite{reed2022a, shridhar2022cliport, shridhar2022perceiver, saycan2022arxiv, driess2023palme, jiang2022vima, brohan2022rt}. These advancements can potentially inspire the development of more autonomous medical robots.

\section{Challenges, Limitations, and Risks}\label{sec:challenges}

Despite the promising outcome of LAMs, there remain many challenges and potential risks in developing and deploying LAMs in biomedical, clinical, and healthcare applications.

\subsubsection{Data}
Most existing public datasets for health informatics are much smaller (please refer to Fig.~\ref{fig:LAM_key_feature_summary} and Table~\ref{tab:LAM_dataset}~\footnote{References to the datasets in Table~\ref{tab:LAM_dataset} and methods in Table~\ref{tab:LAM_sota} can be found in https://github.com/Jianing-Qiu/Awesome-Healthcare-Foundation-Models.}) than those used in general domains and thus are likely insufficient to unlock the full potential of LAMs in biomedical and health scenarios. Building large-scale high-quality medical datasets are particularly challenging because (1) curation requires domain-expertise to identify data of clinical relevance, and quality assurance is very important with health data; (2) some data modalities like MRI require special devices to collect, which is inefficient and expensive; (3) the collected data may not be allowed to publish or use for training because of consent, legal and privacy issues. Furthermore, the training strategy RLHF of some LLMs like ChatGPT requires even more intense engagement of human experts.

\subsubsection{Computation}
Training, or even fine-tuning, contemporary LAMs is extremely expensive in terms of time and resource consumption, which is beyond the budget of most researchers and organizations \cite{ding2023parameter}. Taking LLaMa as an example, an LLM with 65B parameters, it took about 21 days on 2048 A100 GPUs to train the model once on a dataset of 1.4T tokens \cite{touvron2023llama}. Furthermore, even inference can be prohibitively costly due to the model size, making it impractical for most hospitals to deploy these LAMs locally using their computing devices at hand.

\subsubsection{Reliability}
The reliability threshold for translation into clinical practice is significantly higher~\cite{gilbert2023large}. Despite the impressive performance, LLMs are still far from reliable \cite{shen2023chatgpt} and prone to hallucinate \cite{gpt4openai2023, lee2023benefits}, i.e., generating factually-incorrect yet plausible content which misleads users. In addition, the unsatisfactory robustness of LAMs impairs their credibility. LLMs are known to be sensitive to prompts \cite{nori2023capabilities}. LLMs as well as LMs for other modalities remain vulnerable to out-of-distribution and adversarial examples \cite{shen2023chatgpt, wang2023robustness}. Improving the robustness of LAMs may require even more data \cite{li2023data}. Therefore, caution is highly required when using LAMs in healthcare practice to alleviate the potential danger of over-reliance. In addition, LAMs, especially LLMs were trained offline, in many clinical and health scenarios, using up-to-date information is critical.

\subsubsection{Privacy}
First, LAMs have been reported to have excessive capacity to memorize their training data \cite{carlini2021extracting}, and more importantly, it is viable to extract sensitive information in the memorized data using direct prompts \cite{carlini2021extracting, carlini2022quantifying}. This was later mitigated by fine-tuning LAMs to refuse to answer such prompts \cite{li2023multi}. However, Li et al. \cite{li2023multi} also show that this mitigation can be bypassed through tricky prompts called jailbreaking. Moreover, membership inference attacks \cite{shokri2017membership} could reveal if a sample is in the training set, e.g., if a patient is in a cancer dataset. It has been recently demonstrated to work even on the latest large diffusion models \cite{duan2023diffusion}. 

Second, the information provided by users to query LLM-integrated applications may be leaked. According to the data policy of OpenAI \cite{openaidatapolicy2023}, they store the data that users provide to ChatGPT or DALL-E to train their models. Unfortunately, it has been reported that the stored personal information can be leaked incidentally by a ``chat history" bug \cite{openaichatbug2023} or deliberately by indirect prompt injection attack \cite{greshake2023more}.

\subsubsection{Fairness}
LAMs are data-driven approaches so they could learn any bias from the training data. Unfortunately, bias widely exists in the delivery of healthcare \cite{hall2015implicit} and also the data collected in this process \cite{obermeyer2019dissecting, cirillo2020sex, char2018implementing}. Machine learning models trained on such data are reported to mimic human bias against race \cite{obermeyer2019dissecting}, gender \cite{cirillo2020sex}, politics \cite{rutinowski2023self}, etc. In addition to these conventional biases, LLMs present language bias as well, i.e., they perform better in particular languages like English but worse in others \cite{zhuo2023exploring} because training data is dominated by a few languages.

\subsubsection{Toxicity}
Current LAMs, even LLMs explicitly trained with alignment, do not understand and represent human ethics \cite{hendrycks2020aligning}. LLMs are reported to produce hate speech \cite{gehman2020realtoxicityprompts} that causes offensive and psychologically harmful content and even incites violence. Secondly, LAMs may endorse unethical or harmful views and behaviors \cite{hendrycks2020aligning} and motivate users to perform. Lastly, LAMs can be used intentionally to facilitate harmful activities like spreading disinformation and encouraging criminal activities. Although some countermeasures like filtering are applied, they can be circumvented by prompt injection \cite{zhuo2023exploring}.

\subsubsection{Transparency}

Recently, some impactful LAMs like ChatGPT and Med-PaLM 2 chose not to disclose the complete technical details, the pre-trained models, and the used data. This makes it impossible for others to independently reproduce, improve upon and audit their methods. This transparency threat for LAMs can be more serious in healthcare as many medical data is private and models built upon them are not allowed to be open sourced.

\begin{figure}[!t]
\centerline{\includegraphics[width=\columnwidth]{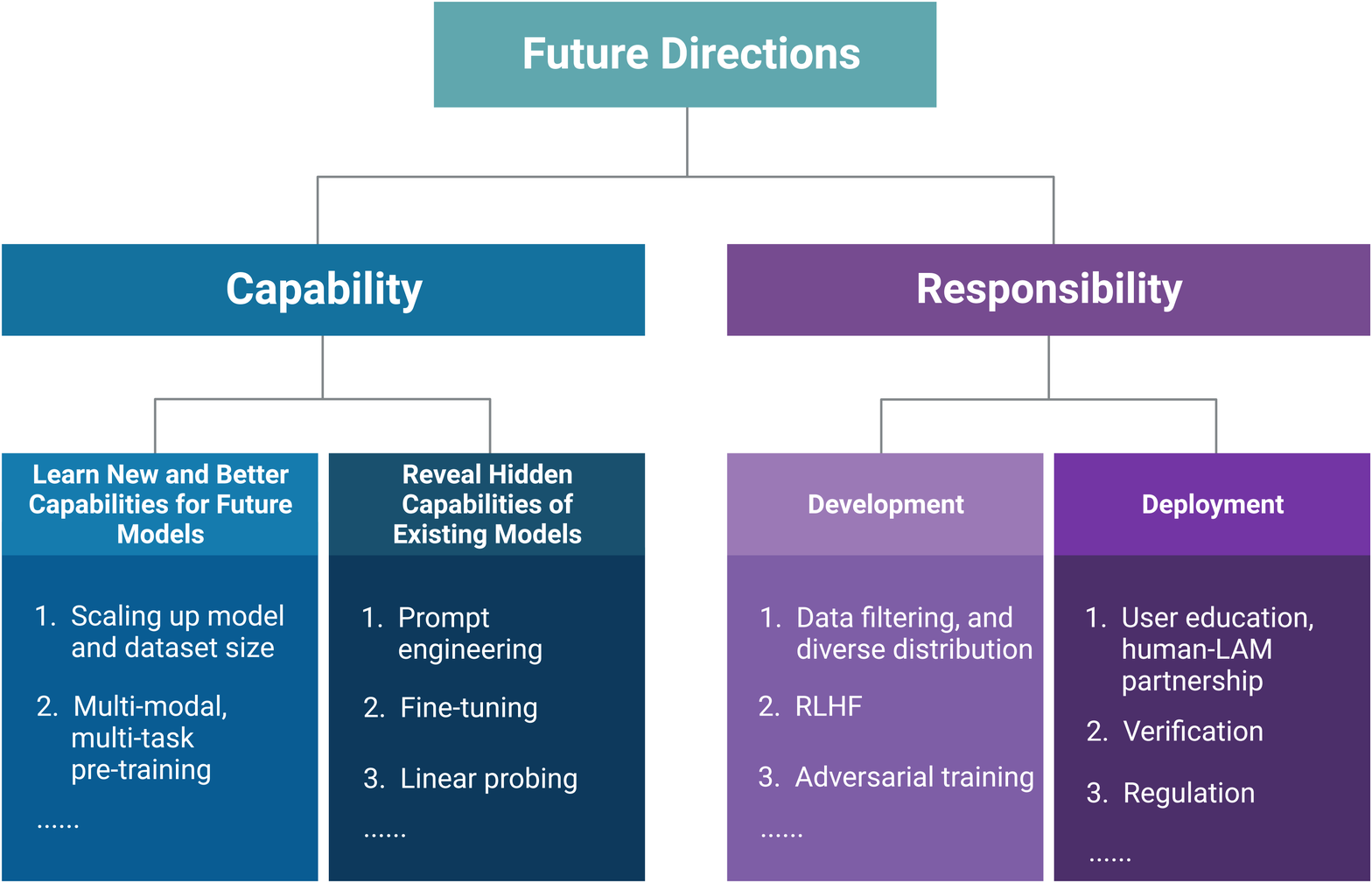}}
\caption{Future directions of LAMs in health informatics}
\label{fig:Future_direction}
\vspace{-0.5cm}
\end{figure}

\subsubsection{Interpretability}
LAMs inherently lack interpretability due to their extremely dense hidden layers. Even worse, the behavior of LAMs can be meaningless \cite{daras2022discovering, wang2023investigating}, hard to predict \cite{kojima2022large, elhage2021mathematical} and thus mysterious. For example, DALL-E 2 generates the images of physical objects with absurd prompts (e.g., ``Apoploe vesrreaitais" for birds) \cite{daras2022discovering}; the reasoning ability of LLMs can be improved by simply adding a text of ``Let’s think step by step" to prompts \cite{kojima2022large}. There has been little progress towards explaining LAMs. Chain-of-thought prompts provide one way to reveal the intermediate reasoning steps behind an output, but it remains unclear whether the generated description of reasoning reflects the model's true internal reasoning. Alternatively, mechanistic interpretability methods \cite{elhage2021mathematical} reverse-engineer the computation of LAMs to illuminate the model's internal mechanism of reasoning.

\subsubsection{Sustainability}
Despite many benefits, LAMs, if abused, will negatively impact the sustainability of our society. LAMs consume lots of computation resource \cite{touvron2023llama} and energy \cite{patterson2021carbon} and emit tons of carbon \cite{patterson2021carbon} in all activities in their lifecycle (from training to deployment) because of their scale. For example, as estimated by \cite{patterson2022carbon}, training a GPT-3 model consumes 1287 MWh and emits 552 tons of \ensuremath{\mathrm{CO_2}}. As the paradigm moves towards LAMs in healthcare, more and more research is expected to be conducted based on LAMs, which could be environmentally unfriendly due to the cost and carbon emission if right practices \cite{patterson2021carbon} are not established.

\subsubsection{Regulation}
Regulation is needed to ensure responsible LAMs especially when some of the above issues cannot be technically addressed. Particularly, data collection and usage should be governed to protect the rights of data owner such as copyright, privacy and ``being forgotten" \cite{villaronga2018humans}. The liability of LAMs' creators/owners for the possible harm caused by the model's output should be clarified. LAMs should be deployed in critical healthcare services only if regulatory approval is obtained and standardized safety assessment is passed. Regulation today is much behind the development of technology for LAMs, even for more general AI. Our webpage lists some major legislation for reference if readers want to know more about AI regulation.

\section{Future Directions}\label{sec:future}

In this section, we discuss some promising directions for future work to advance LAMs in the field of biomedical and health informatics, and our discussion below is mainly focused on two aspects: capability and responsibility.

\subsection{Capability}

The first is to \textbf{develop new LAMs for health informatics with better capability}. The better capability here refers to either new abilities (e.g., a versatile medical task solver) or improved existing abilities (e.g., higher diagnostic accuracy), compared to the prior paradigm. Interestingly, some emergent new abilities may be unexpected or even unknown to humans \cite{wei2022emergent}. Among numerous approaches to LAMs, some are perceived by us as most promising. Scaling up the size of dataset and model are two widely recognized approaches, but how to do it efficiently is of importance and far from solved. Furthermore, pre-training with varied tasks and modalities has achieved remarkable progress towards versatility in performing downstream tasks. A huge benefit is foreseeable if diverse knowledge that exists in these varied tasks (e.g., biology, medicine, etc.) and data modalities (e.g., medical corpora, imaging, physiological signals, etc.) can be incorporated into a single foundation model as a world model \cite{lecun2022path}. This world model boosts capability by complementing the information missing in an input, e.g., offering biomedical knowledge (acquired from other tasks) for diagnosing a disease when only the symptom data is given as input. Note that this is exactly how human doctors diagnose in practice, i.e., they comprehend information from the symptoms based on their medical knowledge acquired from learning and clinical practicing, i.e., multiple other tasks and sources.

The second is to \textbf{reveal the hidden capabilities of existing pre-trained LAMs}. A capability is hidden if it has been already developed in a pre-trained model but just unknown to users. Discovering hidden capabilities involves nothing but probing the model. A typical example is the substantially improved reasoning ability of LLMs by simply adding a line of ``Let's think step by step" to prompts \cite{kojima2022large}. There are still many unknowns about existing LAMs as they have become increasingly complex with enormously large sets of parameters. It is unclear whether the full potential of existing LAMs has been harnessed or not. Therefore, it is worth investigating if those pre-trained LAMs possess hidden capabilities about the health informatics tasks of interest. If so, discovering these hidden capabilities provides a solution or improvement to the tasks in a nearly cost-free way as it requires no further large-scale training. Prompt engineering \cite{liu2023pre} as an emerging field is an effective approach to discovering hidden capabilities.

\subsection{Responsibility}
Responsible LAMs for social good is paramount \cite{leslie2019understanding}. We suggest two complementary strategies: development and deployment, for future work to tackle challenges in LAM reliability, fairness, transparency, and beyond. Development strategy focuses on learning responsible LAMs, while deployment strategy emphasizes using LAMs responsibly.

Technically, responsible LAMs can be developed through two perspectives: data and algorithms. As LAMs learn bias from training data, an intuitive countermeasure is thus to mitigate bias in training data. It can be done by filtering biased data \cite{siddiqui2022metadata}, increasing underrepresented populations' data, etc. Unfortunately, how to efficiently inspect large-scale datasets remains challenging. Besides, training data should encompass diverse distributions to robustify the model against the distribution shifts in the wild, and human preference to align the model with human values. Some of them like human preference must be collected from human activities, while the rest can be also generated by other algorithms like data augmentation and generative models. Once we have high-quality data, algorithms like RLHF and adversarial training can be adopted to exploit these data to acquire the desired properties for responsible LAMs.

In addition to training LAMs to be responsible, it is also vital to use LAMs in a responsible way. Efforts should be made to educate the users, especially those use LAMs for critical healthcare services, about the basics and limitations of LAMs being used. Human-LAM partnership should also be researched for the effective, efficient and responsible use of LAMs, including how to query/instruct LAMs by prompt engineering and assess/adopt the responses from LAMs. Besides, a comprehensive verification framework \cite{huang2023survey} covering various desired properties for LAMs is critical for assessing how irresponsible a LAM is, which is still lacking. We encourage future work to design methods to better evaluate, verify and benchmark LAMs. Last, rules and regulations should be implemented to govern the development, deployment and use of LAMs. This is a vital measure to enforce LAMs for social good and prevent anti-social usage. Overall, building responsible LAMs calls a closer collaboration in the future among academia, industry and government.

\section{Conclusions}\label{sec:conclusions}

We highlight an ongoing paradigm shift within AI community, which is fostering large AI models for transforming different biomedical and health sectors. The new paradigm aims to learn a \textbf{versatile} foundation model on a large-scale (multi-modal) dataset covering \textbf{varied data distributions and learning tasks}. Boundaries between different intelligent tasks, and even between different data modalities, are being dismantled. With generalist intelligence and more unknown capabilities activated, we believe large AI models will augment, instead of replacing, medical professionals and practitioners in the future. Human-AI cooperation will become pervasive. In this regard, the development of large AI models requires even closer and more intense collaboration between domain experts, as well as gradually established regulations.

\bibliographystyle{IEEEtran}
% \bibliography{reference}

\end{document}